\documentclass{article} 
\usepackage{nips15submit_e,times}
\usepackage{graphicx,epsfig}
\usepackage{subfig}
\usepackage{url,cite,amsmath,bm,amssymb}

\newcommand{\mat}[1]{\mathbf{#1}}
\newcommand{\super}[2]{{#1}^{(\text{#2})}}
\newcommand{\vsuper}[2]{\vec{#1}^{(\text{#2})}}
\newcommand{\msuper}[2]{\mat{#1}^{(\text{#2})}}

\newcommand{\supersub}[3]{{#1}^{(\text{#2})}_{\text{#3}}}

\newcommand\vv{\vec{v}}

\newcommand\sizev[1]{\in \mathbb{R}^{#1}}
\newcommand\sizem[2]{\in \mathbb{R}^{#1 \times #2}}

\newcommand\xst{\supersub{x}{s}{t}}
\newcommand\xstu{\supersub{x}{s,u}{t}}

\newcommand\cst{\supersub{c}{s}{t}}
\newcommand\csT{\supersub{c}{s}{T}}
\newcommand\hst{\supersub{h}{s}{t}}
\newcommand\hsT{\supersub{h}{s}{T}}
\newcommand\hstm{\supersub{h}{s}{t-1}}

\newcommand\hik{\super{h}{i,k}}
\newcommand\hikm{\super{h}{i,k-1}}

\newcommand\htt{\supersub{h}{t}{j}}
\newcommand\htT{\supersub{h}{t,k-1}{T}}
\newcommand\httm{\supersub{h}{t}{j-1}}
\newcommand\ytt{\supersub{y}{t}{j}}
\newcommand\yttu{\supersub{y}{t,u}{j}}
\newcommand\yttm{\supersub{y}{t}{j-1}}
\newcommand\yttstt{\supersub{y}{t}{1}}
\newcommand\ctt{\supersub{c}{t}{j}}
\newcommand\vxs{\vsuper{x}{s}}

\newcommand\vxsu{\vsuper{x}{s,u}}
\newcommand\vytu{\vsuper{y}{t,u}}
\newcommand\Wae{\msuper{W}{ae}}

\newcommand\Wah{\msuper{W}{ah}}

\title{Attention with Intention for a Neural Network Conversation Model}
\author{
Kaisheng Yao~\thanks{Presented at NIPS Workshop on Machine Learning for Spoken Language Understanding and Interaction 2015.}\\
Microsoft Research\\
\texttt{kaisheny@microsoft.com} 
\and
Geoffrey Zweig\\
Microsoft Research\\
\texttt{gzweig@microsoft.com} 
\and
Baolin Peng\\
Chinese University of Hong Kong\\
\texttt{blpeng@se.cuhk.edu.hk} 
}

\nipsfinalcopy 

\begin{document}
\maketitle

\begin{abstract}
In a conversation or a dialogue process, attention and intention play intrinsic roles. This paper proposes a neural network based approach that models the attention and intention processes. It essentially consists of three recurrent networks. The encoder network is a word-level model representing source side sentences. The intention network is a recurrent network that models the dynamics of the intention process. The decoder network is a recurrent network that produces responses to the input from the source side. It is a language model that is dependent on the intention and has an attention mechanism to attend to particular source side words, when predicting a symbol in the response. The model is trained end-to-end without labeling data. Experiments show that this model generates natural responses to user inputs. 
\end{abstract}

\section{Introduction}

A conversation process is a process of communication of thoughts through words. It may be considered as a structural process that stresses the role of purpose and processing in discourse~\cite{Grosz1986Discourse}. Essentially, the discourse structure is intimately connected with two nonlinguistic notions: intention and attention. In  processing an utterance, attention explicates the processing of utterances, for example, paying attention to particular words in a sentence. On the other hand, intention is higher level than attention and has its primary role of explaining discourse structure and coherence. 
Clearly, a conversation process is inherently complicated because of the two levels of  structures. 

A conversation process may be cast as a sequence-to-sequence mapping task. In this task, the source side of the conversation is from one person and the target side of the conversation is from another person. The sequence-to-sequence mapping task includes machine translation, grapheme-to-phoneme conversion, named entity tagging, etc. However, an apparent difference of a dialogue process from these tasks is that a dialogue process involves multiple turns, whereas usually the above tasks involve only one turn of mapping a source sequence to its target sequence. 

Neural network based approaches have been successfully applied in sequence-to-sequence mapping tasks. They have made significant progresses in machine translation~\cite{sutskever2014sequence,bahdanau2015neural,DevlinNNJSMT}, language understanding~\cite{Mesnil2015RNNLU}, and speech recognition~\cite{Chan2015LAS}.
Among those neural network-based approaches, one particular approach, which is called encoder-decoder framework~\cite{sutskever2014sequence,bahdanau2015neural}, aims at relaxing much requirement on human labeling. 


Conversation models have been typically designed to be domain specific with much knowledge such as rules~\cite{Bohus2009Dialogue,Young2013POMDP}. Recent methods~\cite{Wen2015SLG} relax such requirement to some extent but their whole systems are still trained with manual labels because of their sub-components that require so. 
Manual labels are error prone and expensive. Therefore, it is appealing to train a system end-to-end without manual labels. Recent works in \cite{Vinyals2015NeuralConversationModel,Sordoni2015Conversation,Shang2015NRM} are in this approach. 

In general, however, using knowledge is helpful. For example, the alignment information between the source and target side is critical in grapheme-to-phoneme conversion~\cite{Yao2015S2SG2P} to outperform a strong baseline using n-gram models~\cite{Bisani2008G2P}. In a neural network based machine translation system~\cite{DevlinNNJSMT}, the alignment information is used to outperform a strong phrase-based baseline~\cite{Chiang2007Hiero}. 

In the context of modeling conversation process, a neural network model may be built with the knowledge of the structural information of conversation processes. In particular, the network may incorporate the notion of intention and attention. To test this, we developed a model that consists of three recurrent neural networks (RNNs). The source side RNN, or encoder network, encodes the source side inputs. The target side RNN, or decoder network, uses an attention mechanism to attend to particular words in the source side, when predicting a symbol in its response to the source side. Importantly, this attention in the target side is conditioned on the output from an intention RNN. This model, which has the structural knowledge of the conversation process, is trained end-to-end without labels. We experimented with this model and observed that it generates natural responses to user inputs.

\section{Background}
In the theory of discourse in \cite{Grosz1986Discourse}, discourse structure is composed of three separate but related components. The first is the linguistic structure, which is the structure of the sequence of utterance. The linguistic structure consists of segments of the discourse into which the utterances naturally aggregate. The second structure is the intentional structure, which captures the discourse-relevant purposes, expressed in each of the linguistic segments as well as relationships among them. 
The third is the attentional state that is dynamic, and records the objects, properties, and relations that are salient at each point of the discourse. 

In many examples we observe, there are usually just one linguistic segment that consists of all the utterances. Therefore, in the following, we consider a discourse with two structures: intention and attention. 

In the example in Table~\ref{tab:exp}, there is a clear flow of intentions. The user states the problem, with the user's intention of conveying the problem to the agent. The agent receives the words, processes them, and communicates back to the user. The user responds to the agent afterwards. Therefore, the whole conversation process consists of three intentions processed sequentially. The first is the intention of communication of the problem. The second intention is the process of resolving the issue. The third is the intention of acknowledgment. In processing each of the intentions, the user and the agent pay attention to particular words. For example, when resolving the issue, the agent pays attention to words such as "virus". 

\begin{table}[t]
\caption{An example of dialogue process. \label{tab:exp}}
\begin{center}
\begin{tabular}{c|l}
\hline
\textit{user}&  my computer is infected \\
\textit{agent} & do you want to retrieve the files that was deleted? \\
\textit{user}& the ones that the virus deleted , yes. \\
\textit{agent} & i can help you resolve the issue with our virus removal and protection service \\
\textit{user}& ok \\
\textit{agent}&  here is a link how to run system restore \\
\textit{user}&  thank you . \\
\textit{agent}& you welcome \\
\hline
\end{tabular}
\end{center}
\end{table}

\section{The model}
\subsection{The attention with intention (AWI) model}
\begin{figure}[t]
\centering
\includegraphics[width=0.65\columnwidth]{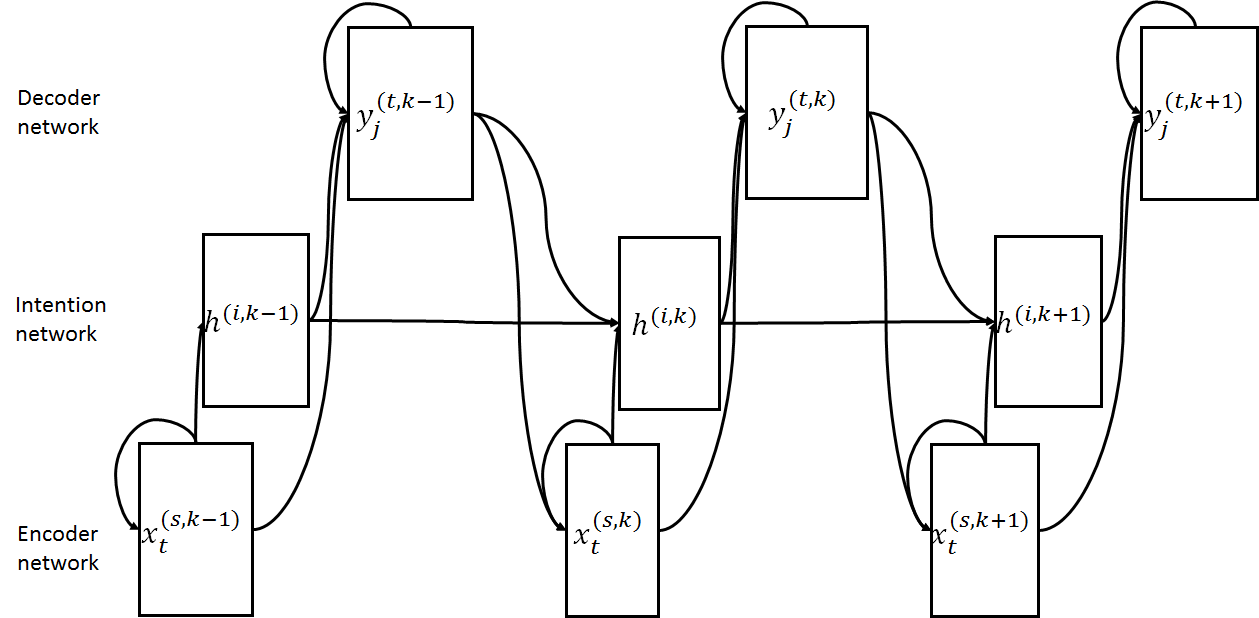}
\caption{The attention with intention (AWI) model. The model is unrolled into three turns. In each turn has RNNs for encoder network and decoder network. Each session is represented by a fixed-dimension vector, which is a hidden state of an intention RNN network. }
\label{fig:awi}
\end{figure}

We propose a model that attempts to represent the structural process of intentions and the associated attentions. Figure~\ref{fig:awi} illustrates the model. It shows three layers of processing: encoder network, intention network, and decoder network. 

The encoder network has inputs from the current source side input. Because the source side in the current turn is also dependent on the previous turn, the source side encoder network is linked with the output from the previous target side. The encoder network creates a representation of the source side in the current turn. 

The intention network is dependent on its past state, so that it memories the history of intentions. It therefore is a recurrent network, taking a representation of the source side in the current turn and updating its hidden state.

The decoder is a recurrent network for language modeling that outputs symbol at each time. This output is dependent on the current intention from the intention network. It also pays attention to particular words in the source side. 

In more details, a conversation has in totoal $U$ turns. At turn $u$, a user in the source side, denoted in superscript $^{(s)}$, has an input sequence of $\vxsu = (\xstu : t = 1, \cdots, T)$ with length $T$. An agent in the target side, denoted in superscript $^{(t)}$, responds to the user with $\vytu = (\yttu : j = 1, \cdots J)$ with length $J$. The proposed model is a conditional model of the target
given the source, $p(\vytu | \vxsu)$. 
If there is no confusion, we may omit the session index $u$ in the following.

\subsection{Encoder network}
\label{sec:encoder}
The encoder network reads the input sentence $\vxs$, and converts them into a fixed-length or a variable length representation of the source side sequence. There are many choices to encode the source side. The approach we use is an RNN such that 
\begin{equation}
\hst = f\left(\xst , \hstm\right)
\end{equation}
\noindent where $f(\cdot)$ is an RNN. $\hst$ is the hidden state at time $t$ in the source side. 
The initial state $\hst$ with $t=0$ is the last hidden state activity, $\htT$, of the decoder network in the previous turn $k-1$. 

One form of the output from this encoder is the last hidden state activity $\csT = \hsT$. This is used as a representation of the source side in the current turn to the intention network. 
The other form is a variable-length representation, to be used in the attention model described in Sec.~\ref{sec:decoder}. A general description of the variable length representation is as follows
\begin{equation}
\cst = q(\{\hst, \forall t=0,\cdots T\})
\end{equation}
\noindent where $q(\cdot)$ might be a linear network or a nonlinear network. 

\subsection{Intention network}
\label{sec:intention}
The signal from the encoder network is fed into an intention network to model the intention process. 
Following \cite{Grosz1986Discourse}, the intention process is a dynamic process to model the intrinsic dynamics of conversation, in which an intention in one turn is dependent on the intention in the previous turn. This property might be modeled using a Markov model, but we choose an RNN. 

Interestingly, the hidden state of an RNN in a certain turn may be considered as a distributed representation of the intention. Different from the usual process of training distributed representation of words~\cite{Mikolov2013Word2Vec}, the distribution representation of intentions are trained with previous turns as their context. We use a first order RNN model, in which a hidden state is dependent explicitly on its previous state. 

The intention model in AWI is therefore an RNN as follows
\begin{equation}
\hik = f\left( \csT, \hikm, \htT \right)
\end{equation}
\noindent where $\csT$ is the fixed dimension representation of the source side described in Sec.~\ref{sec:encoder}. $k$ is the index of the current turn. $\htT$ is the last hidden layer activity of the decoder network in the previous turn $k-1$.

\subsection{Decoder network}
\label{sec:decoder}
The last step is to decode the sequence in the target side, which
is framed as a language model over each symbol, generated left to
right. 
In this framework, the decoder computes conditional probability as 
\begin{equation}
p(\ytt|\yttstt,\cdots,\yttm,\vxs)  = g(\yttm, \htt, \ctt)
\end{equation}
\noindent where the hidden state in the decoder is computed using an RNN
\begin{align}
\htt = f \left( \yttm, \httm, \ctt\right)
\end{align}
The initial state $\htt$ at $t=0$ is the last hidden state activity from the intention network. 

The $\ctt$ is a vector to represent the context to generate $\ytt$. It is dependent on the source side as 
\begin{eqnarray}
\ctt & = & z \left(\httm, \{\cst: \forall t=\{1,\cdots,T\}\}\right) 
\end{eqnarray}
\noindent where $z(\cdot)$ summerizes the variable-length source side representations $\{\cst\}$ using weighted average. The weight is computed using a content-based alignment model~\cite{bahdanau2015neural} that produces high scores if the target side hidden state in previous time $\httm$ and $\cst$ are similar. More formally, the weight $\alpha_{jt}$ for the context $\cst$ is computed using
\begin{equation}
\alpha_{jt} = \frac{\exp{e_{jt}}}{\sum_m \exp({e_{jm}})}
\end{equation}
\noindent where 
\begin{equation}
e_{jt} = a(\httm, \cst) \label{eqn:alignment}
\end{equation}

The alignment model enables an attention to particular words, represented as a vector $\cst$ in the source side. Since the decoder network generates responses on condition of the attention and also the intention, our model is called attention with intention (AWI) model. 

\subsection{Implementation details}
All of the recurrent networks are implemented using a recently proposed depth-gated long-short-term memory (LSTM) network~\cite{Yao2015DGLSTM}. The context vector $\cst$ is an embedding vector of the source side word at time $t$.

The alignment model in Eq. (\ref{eqn:alignment}) follows the attention model in \cite{bahdanau2015neural}, in which $e_{jt}$ is calculated as 
\begin{equation}
e_{jt}  = \vv^\top \tanh \left( \Wah \httm + \Wae \cst\right), \label{eq:alignment2} 
\end{equation}
\noindent which is a neural
network with one hidden layer of size $A$ and a single output, parameterised by
$\Wae \sizem{A}{H}$,
$\Wah \sizem{A}{H}$ and
$\vv \sizev{A}$.
$H$ and $A$ are the hidden layer dimension and alignment dimension.

\section{Evaluation}
\label{sec:experiments}
We used an in-house dialogue dataset. The dataset consists of dialogues from a helpdesk chat service. In this service, costumers seeks helps on computer related issues from human agents. Training consists of 10000 dialogues with 96913 turns or conversations. Number of tokens is 2215047 in the source side and 2378950 in the target side. The vocabulary size is 9085 including words from both side. 
Development set data has 1000 dialogues with 9971 turns. 
Test set data has 500 dialogues with 5232 turns. 

We use sentence-level SGD without momentum. Learning rate is initialized to 0.1. Development set is used to control the learning rate. The learning rate is halved when perplexity on the development is increased. 
One epoch of training has one pass of the training data. The order of training dialogues is randomly shuffled in the beginning of each epoch. The order of turns in the dialogue is however kept. 

\subsection{Performances measure in perplexity}
\label{exp:ppl}
An objective comparison of different models for conversation is still an open question. We report perplexity (PPL), though it may have drawbacks, to compare different models. Table~\ref{tab:awiresults} presents results in perplexity with two models with different hiden layer sizes. Results show that a larger model with 200 hidden layer dimension has lower PPL than the model with 50 dimension.

\begin{table}[t]
\caption{Perplexity results with AWI model. Models have 50 or 200 hidden layer dimension, 25 alignment dimension, and one layer of LSTMs. \label{tab:awiresults}} 
\begin{center}
\begin{tabular}{|c|c|}
\hline
\# hidden dimension & PPL  \\
\hline
50  & 30.8 \\
200 & 22.1 \\
\hline
\end{tabular}
\end{center}
\end{table}



\subsection{Examples of outputs from the trained model}
Table~\ref{tab:exptrainedmodel} lists an example of the conversation process between a human and the trained model. The model has two layers of LSTMs and other setups are the same as used in Sec~\ref{exp:ppl}. Similarly as observed in \cite{Sordoni2015HRED}, the model produces natural responses to user inputs. The flow of intentions is clearly seen in this example. 
\begin{table}[ht]
\caption{The generated response from the trained model.\label{tab:exptrainedmodel}}
\begin{center}
\begin{tabular}{c|l}
\hline
\textit{user} & hi \\
\textit{agent} &    how may i help you today ? \\
\textit{user}& my computer is not responding \\
\textit{agent} &    oh ! i sorry to hear that . are you getting an error message ? \\
\textit{user}& yes \\
\textit{agent} &    what recent changes did you made with this computer ? \\
\textit{user}& i updated my computer \\
\textit{agent} &    i see . is this on windows 7 ? \\
\textit{user}& no it is on linux \\
\textit{agent} &    ok . are you using the affected computer right now ? \\
\textit{user}& no \\
\textit{agent} &    i can definitely assist you on your issue . \\
\textit{user}& thanks \\
\textit{agent} &    we can go ahead and start working on this issue within our office premium software support service . \\
\textit{user}& that is allright \\
\textit{agent} &    if in case the issue comes back at all within the next 30 days we will perform the service again for free . \\
\textit{agent} & let me go ahead and prepare the offer and let start working on this issue now , alright ? \\
\textit{user}& please \\
\textit{agent} &    sure . may i have your phone number please . \\
\textit{user}& xxx-xxx-xxxx \\
\textit{agent} &    thank you . \\
\textit{user}& no problem \\
\textit{agent} &    thank you for waiting . \\
\textit{user}& no problem \\
\textit{agent} &    i see . \\
\textit{user}& what is the next step \\
\textit{agent} &    i see . let me go ahead and prepare the offers for you . \\
\textit{user}& okay \\
\textit{agent} &    thank you . \\
\textit{user}& no problem \\
\textit{agent} &    thank you for contacting answer desk . have a great day ! \\
\textit{user}& bye \\
\textit{agent} &    thanks again for working with me . id like to send you a quick survey . \\
\textit{agent} & it starts with a yes/no question , with a few more questions about your experience with me today. \\
\textit{agent} & the survey takes about a minute and really helps us improve our service . can i send it to you now ? \\
\textit{user}& sure \\
\textit{agent} &    thank you . \\
\textit{user}& no problem \\
\textit{agent} &    thank you for contacting answer desk . have a great day ! \\
\textit{user}& bye \\
\textit{agent} &    goodbye . \\
\hline
\end{tabular}
\end{center}
\end{table}

\section{Related work}
\label{sec:related}
Our work is related to the recent works in~\cite{Shang2015NRM,Vinyals2015NeuralConversationModel,Sordoni2015Conversation}, which use an encoder-decoder framework to model conversation. The work in \cite{Shang2015NRM} is a model for single turn conversation. The work in \cite{Vinyals2015NeuralConversationModel} is a simple encoder-decoder method using a fixed-dimension representation of the source side. The work in \cite{Sordoni2015Conversation} also uses a fixed-dimension representaiton of the source side but has an additional RNN to model dialogue context. This additional RNN is similar to the intention RNN in AWI model. 

However, AWI model differs from \cite{Sordoni2015Conversation} in that it incorprates the concept of attention and intention based on the theory in \cite{Grosz1986Discourse}. Therefore, attention mechanism is essential to AWI. The model in \cite{Sordoni2015Conversation} doesn't have an attention model. 

Because it is not yet clear what objective measure to use to compare different models, it is hard to make claims of superiority of these models. We believe AWI model is an alternative to the models in \cite{Vinyals2015NeuralConversationModel,Sordoni2015Conversation}. 

\section{Conclusions and discussions}
We have presented a model that incorporates attention and intention processes in a neural network model. Preliminary experiments show that this model generates natural responses to user inputs. Future works include experiments on common dataset to compare different models and incorporating objective functions such as goals. 
 
\bibliographystyle{abbrv}
\bibliography{cite-strings,../ref/refs,cite-definitions}

\begin{thebibliography}{10}

\bibitem{bahdanau2015neural}
D.~Bahdanau, K.~Cho, and Y.~Bengio.
\newblock Neural machine translation by jointly learning to align and
  translate.
\newblock In {\em {Proceedings of the International Conference on Learning
  Representations (ICLR)}}, San Diego, CA, 2015.

\bibitem{Bisani2008G2P}
M.~Bisani and H.~Ney.
\newblock Joint-sequence models for grapheme-to-phoneme conversion.
\newblock {\em Speech Communication}, 50(5), 2008.

\bibitem{Bohus2009Dialogue}
D.~Bohus and A.~I. Rudnicky.
\newblock The ravenclaw dialog management framework: architecture and systems.
\newblock {\em Computer, Speech and Language}, 23:332--361, 2009.

\bibitem{Chan2015LAS}
W.~Chan, N.~Jaitly, Q.~V. Le, and O.~Vinyals.
\newblock Listen, attend and spell.
\newblock In {\em arXiv:1508.01211 [cs.CL]}, 2015.

\bibitem{Chiang2007Hiero}
D.~Chiang.
\newblock Hierarchical phrase-based translation.
\newblock {\em Computational Linguistics}, 33(2):201--228, 1999.

\bibitem{DevlinNNJSMT}
J.~Devlin, R.~Zbib, Z.~Huang, T.~Lamar, R.~Schwartz, and J.~Makhoul.
\newblock Fast and robust neural network joint models for statistical machine
  translation.
\newblock In {\em ACL}, 2014.

\bibitem{Grosz1986Discourse}
B.~J. Grosz and C.~L. Sidner.
\newblock Attention, intentions, and the structure of discourse.
\newblock {\em Computational Linguistics}, 12:175--204, 1986.

\bibitem{Mesnil2015RNNLU}
G.~Mesnil, Y.~Dauphin, K.~Yao, Y.~Bengio, L.~Deng, D.~Hakkani-Tur, X.~He,
  L.~Heck, G.~Tur, D.~Yu, and G.~Zweig.
\newblock Using recurrent neural networks for slot filling in spoken language
  understanding.
\newblock {\em IEEE/ACM Transactions on Audio, Speech, and Language
  Processing}, 2015.

\bibitem{Mikolov2013Word2Vec}
T.~Mikolov, K.~Chen, G.~Corrado, and J.~Dean.
\newblock Effcient estimation of word representations in vector space.
\newblock In {\em NIPS}, 2013.

\bibitem{Shang2015NRM}
L.~Shang, Z.~Lu, and H.~Li.
\newblock Neural responding machine for short-text conversation.
\newblock In {\em ACL}, 2015.

\bibitem{Sordoni2015HRED}
A.~Sordoni, Y.~Bengio, H.~Vahabi, C.~Lioma, J.~G. Simonsen, and J.-Y. Nie.
\newblock A hierarchical recurrent encoder-decoder for generative context-aware
  query suggestion.
\newblock In {\em arXiv:1507--0222 [cd.NE]}, July 2015.

\bibitem{Sordoni2015Conversation}
A.~Sordoni, M.~Galley, M.~Auli, C.~Brockett, Y.~Ji, M.~Mitchell, J.-Y. Nie,
  J.~Gao, and B.~Dolan.
\newblock A neural network approach to context-sensitive generation of
  conversation responses.
\newblock In {\em NAACL}, 2015.

\bibitem{sutskever2014sequence}
I.~Sutskever, O.~Vinyals, and Q.~V. Le.
\newblock Sequence to sequence learning with neural networks.
\newblock In {\em {Neural Information Processing Systems (NIPS)}}, pages
  3104--3112, Montr\'eal, 2014.

\bibitem{Vinyals2015NeuralConversationModel}
O.~Vinyals and Q.~V. Le.
\newblock A nerual converstion model.
\newblock In {\em ICML Deep Learning Workshop}, 2015.

\bibitem{Wen2015SLG}
T.-H. Wen, M.~Gasic, D.~Kim, N.~Mrksic, P.-H. Su, D.~Vandyke, and S.~Young.
\newblock Stochastic language generation in dialogue using recurrent neural
  networks with convolutional sentence reranking.
\newblock Technical report, May 2015.

\bibitem{Yao2015DGLSTM}
K.~Yao, T.~Cohn, E.~Vylomova, K.~Duh, and C.~Dyer.
\newblock Depth-gated \textup{LSTM}.
\newblock In {\em arXiv:1508.03790 [cs.NE]}, 2015.

\bibitem{Yao2015S2SG2P}
K.~Yao and G.~Zweig.
\newblock Sequence-to-sequence neural net models for grapheme-to-phoneme
  conversion.
\newblock In {\em INTERSPEECH}, 2015.

\bibitem{Young2013POMDP}
S.~Young, M.~Gasic, B.~Thomson, and J.~D. Williams.
\newblock \textup{POMDP}-based statistical spoken dialog systems: A review.
\newblock {\em Proceedings of the \textup{IEEE}}, 101:1160--1179, 2013.

\end{thebibliography}

\end{document}